\documentclass[10pt,twocolumn,letterpaper]{article}
\usepackage[pagenumbers]{cvpr} 

\usepackage[dvipsnames]{xcolor}

\newcommand{\R}{\mathbb{R}}

\newcommand{\model}{\epsilon_\theta}
\newcommand{\conditioner}{\tau_\theta}

\definecolor{cvprblue}{rgb}{0.21,0.49,0.74}
\usepackage[pagebackref,breaklinks,colorlinks,citecolor=cvprblue]{hyperref}
\usepackage{verbatim}
\usepackage{amsmath}
\usepackage{graphicx}
\usepackage{multirow}
\usepackage{overpic}
\usepackage[linesnumbered, boxed, ruled]{algorithm2e}
\begin{document}

\title{Cut-and-Paste: Subject-Driven Video Editing with Attention Control}

\author{
    Zhichao Zuo\textsuperscript{1},
    Zhao Zhang\textsuperscript{1*},
    Yan Luo\textsuperscript{1},
    Yang Zhao\textsuperscript{1},
    Haijun Zhang\textsuperscript{2},
    Yi Yang\textsuperscript{3},
    Meng Wang\textsuperscript{1}\\
{\small \textsuperscript{1}School of Computer Science and Information Engineering, Hefei University of Technology, Hefei, China}, \\ {\small \textsuperscript{2}School of Computer Science, Harbin Institute of Technology, Shenzhen, China} \\{\small \textsuperscript{3}School of Computer Science and Technology, ZhejiangUniversity, Hangzhou, China}
}

\twocolumn[{
\renewcommand\twocolumn[1][]{#1}
\maketitle
\begin{center}
    \centering
    \vspace*{-.8cm}
    \includegraphics[width=\textwidth]{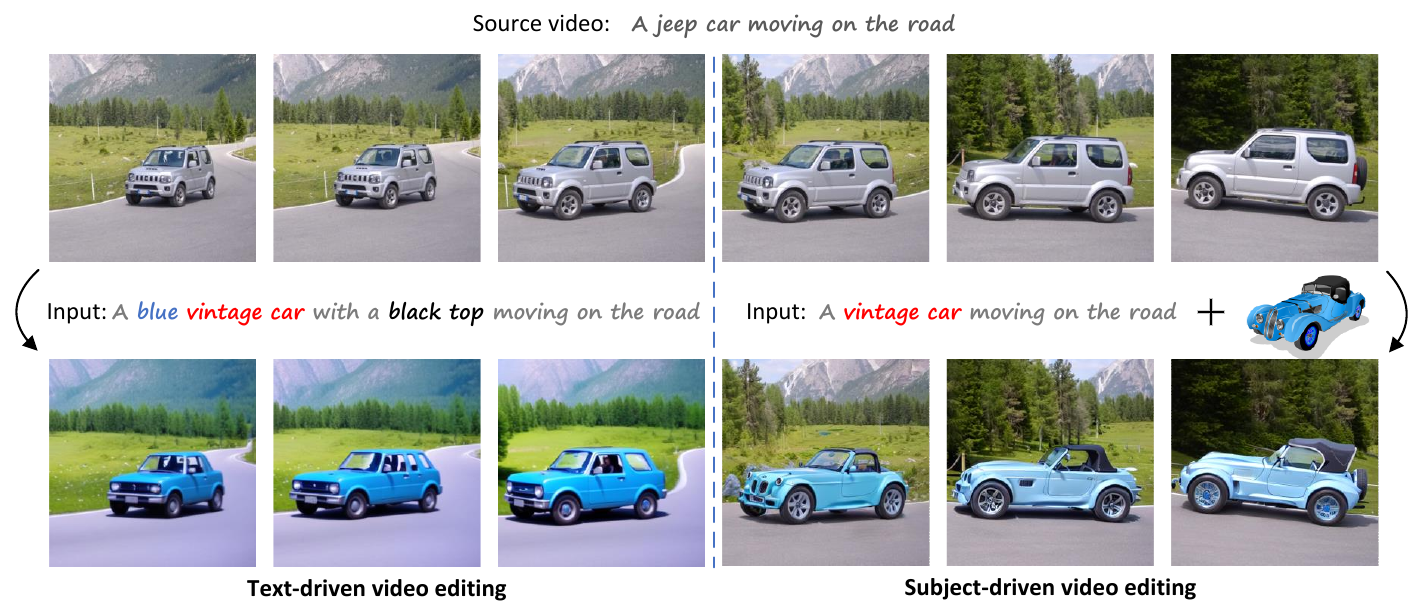}
    \vspace*{-.6cm}
    \captionof{figure}{\textbf{Text-driven video editing} vs. \textbf{Subject-driven video editing} (ours). For precise control of edited contents, text-driven video editing methods require a cumbersome text input to describe various aspects of the object's properties. Nevertheless, the results are always unsatisfactory and the background usually changes as well \textbf{(Left)}. In contrast, we propose a novel framework termed \textit{Cut-and-Paste} for subject-driven video editing, which leverages the reference image as supplementary input and then just needs a simple text prompt, which achieves fine-grained semantic generation and meanwhile preserves the background of source video better with attention control \textbf{(Right)}.}
\label{fig:cover}
\end{center}
}]
\begin{abstract}
This paper presents a novel framework termed Cut-and-Paste for real-word semantic video editing under the guidance of text prompt and additional reference image. While the text-driven video editing has demonstrated remarkable ability to generate highly diverse videos following given text prompts, the fine-grained semantic edits are hard to control by plain textual prompt only in terms of object details and edited region, and cumbersome long text descriptions are usually needed for the task. We therefore investigate subject-driven video editing for more precise control of both edited regions and background preservation, and fine-grained semantic generation. We achieve this goal by introducing an reference image as supplementary input to the text-driven video editing, which avoids racking your brain to come up with a cumbersome text prompt describing the detailed appearance of the object. To limit the editing area, we refer to a method of cross attention control in image editing and successfully extend it to video editing by fusing the attention map of adjacent frames, which strikes a balance between maintaining video background and spatio-temporal consistency. Compared with current methods, the whole process of our method is like ``cut" the source object to be edited and then ``paste" the target object provided by reference image. We demonstrate that our method performs favorably over prior arts for video editing under the guidance of text prompt and extra reference image, as measured by both quantitative and subjective evaluations.
\end{abstract}    
\section{Introduction}
\label{sec:intro}

Text-driven video editing, which enables versatile and high-quality translations of the original video content only by textual prompts, has flourished with the development of large-scale text-to-image (T2I) generative diffusion models~\cite{ldm,glide,classifierfree,clip,ddim,ddpm}. The strong semantic prior learned from a large collection of image-caption pairs is one of the most important reasons that enable the task to be accomplished. For instance, when presenting a video with a textual description of ``a jeep car moving on the road”, we can transform the original video to the edited version of ``a vintage car moving on the road" , just by changing the words ``jeep car" to ``vintage car". 

Predictably, to enable more fine-grained editing and more precise control, users need to provide more detailed and standardized textual descriptions (\eg detailed appearance of the vintage car, see Figure~\ref{fig:cover} left). However, the neural and inherent gap between vision and language makes it not as easy as it should be, the semantic prior learned from the T2I models is not sufficient to support fine-grained semantic control. Tremendous efforts~\cite{PromptEngineering,ipl} have been made to improve this issue. Prompt-engineering~\cite{PromptEngineering,ipl}, the art of writing text prompts to get an AI system to generate the output you want, serves as a lite alternative for personalizing T2I models. Nevertheless, searching high-quality text prompts for customized results is more art than science. And even detailed textual descriptions inevitably lead to ambiguity and may not accurately reflect the desired effects of users. In fact, many details of object appearance are challenging to convey through ordinary language only. Such a problem also exists in the area of text-driven image editing, which tends to manipulate image just guided by text prompt. Presently personalized tuning techniques~\cite{dreambooth,paintbyexample,plugandplay,promptfreediffusion,controlnet}, such as Dreambooth~\cite{dreambooth} and ControlNet~\cite{controlnet}, have provided great ideas for solving the above problems by using additional images as supplementary input. These methods present a fresh perspective for tackling this issue, although it may be challenging to be applied directly to text-driven video editing.

In this paper, we therefore define and propose a subject-driven video editing approach, manipulating the video under the guidance of both text prompt and a reference image, which allows more accurate semantic manipulation and provides more precise control on the video content. As commonly argued: ``an image is worth a thousand words", we believe that an image can express the editing effect the user wants better than text descriptions. Thus, we introduce an additional reference image that corresponds to the target edited object to provide fine-grained semantic information. In this way, the cumbersome textual description of the edited object is essentially replaced by the features embedded in the reference image, what we need is just a simple text prompt and a reference image (see Figure~\ref{fig:cover} right). 

Another more challenging scenario of text-driven video editing, on the other hand, is the precise control ability to the target editing area. When we try to edit just one subject in the video, we need to change the target subject word in the text prompt. However, due to the instability of T2I pre-trained diffusion models and the consistency across frames, previous works on text-driven video editing are hard to control the editing region effectively, often leading to unexpected changes in the non-editing region. Even a small change of the text prompt often causes a completely different outcome, hence challenging to preserve the structure and composition of the original video. In order to preserve the layout and semantics in unedited regions, prior methods~\cite{blendeddiffusion,paintbyexample} adopt a strategy of manually marking editing areas, which only modifies the masked area without affecting the unmasked regions. However, manually masking each frame of a video is obviously time-consuming and cumbersome. A wonderful method to control editing area in image editing is Prompt-to-Prompt (P2P)~\cite{prompttoprompt}, which automatically localizes the editing region by manipulating the cross-attention layers, but it is challenging to be applied directly to semantic video editing owing to the spatio-temporal consistency across video frames. Since visual and textual embedding are fused using cross-attention layers that produce spatial attention maps for each textual token, we propose to inject the attention maps of the previous frame to the current frame before performing the Word Swap operation in P2P~\cite{prompttoprompt} in some steps of the diffusion process, which effectively strikes a balance between maintaining the video structure and  spatio-temporal consistency across frames.

In summary, our contributions are described as follows:
\begin{itemize}
    \item We define and come up with a new video editing method, termed subject-driven video editing, which adds an extra reference image to the input of text-driven video editing for more precise and fine-grained semantic control.
    \item We propose attention control with adjacent frames, which effectively strikes a balance between maintaining video structure and spatio-temporal consistency across frames.
    \item Extensive quantitative and numerical experiments have demonstrated the remarkable editing ability of our approach and establish its superior performance compared to general text-driven video editing.
\end{itemize}
\section{Related Work}
\label{sec:ralated}
\subsection{Image Editing with Diffusion Model}
Existing image editing methods based on diffusion model can be mainly divided into two categories, \ie, \textit{text-driven} and \textit{subject-driven}. Text-driven image editing ~\cite{imagic,plugandplay,prompttoprompt,unitune} tend to manipulate image only using text as prompt. The advent and rapid rise of diffusion models~\cite{ldm,ddpm,ddim,classifierfree,clip,glide,wang2022fineformer,Wang2023CropCap,zhao2022fclgan} makes it possible to generate high-quality and vivid editorial contents for text-driven image editing. Both Prompt-to-Prompt~\cite{prompttoprompt} and Plug-and-Play~\cite{plugandplay} enable diverse translations of the original image content through making changes to the text prompt entered. UniTune~\cite{unitune} and Imagic~\cite{imagic} also achieve amazing results while maintaining good fidelity by fine-tuning on a single image. Blended Diffusion~\cite{blendeddiffusion} modifies the masked area according to a guiding text prompt along with an ROI mask. These methods work well in specified scenarios and often yield realistic editing results. While limited to the gap of text and image, most of the text-driven image editing approaches can only edit images primarily on specific domains instead of open-world text sets. Then the frameworks~\cite{textualinversion,dreambooth,paintbyexample,blipdiffusion} of subject-driven image editing, for which additional images are introduced as references, emerged and achieved better editing results in terms of control ability. DreamBooth~\cite{dreambooth} can generate a myriad of images of the subject in different contexts with just a few images of a subject, using the guidance of a text prompt. Similar to~\cite{blendeddiffusion}, Paint by Example~\cite{paintbyexample} also introduce an arbitrary shape mask and semantically alters the image content of the masked area, under the guidance of both text and image. 

The aforementioned studies have achieved excellent results in image editing, nevertheless, directly applying these methods to each frame independently for video editing often leads to temporal inconsistencies and cannot achieve satisfactory results. Thus, we introduce an efficient tuning strategy that only updates the projection matrices in attention blocks like~\cite{fatezero} and propose the attention control with adjacent frames, which strikes a balance between preserving video structure and spatio-temporal consistency.
\subsection{Video Editing with Diffusion Model}
In contrast to the image editing, video editing task is significantly more challenging in terms of the generated outcome on a single frame and the temporal disparity of adjacent frames. Recent works~\cite{text2live,tuneavideo,fatezero,dreamix} have made considerable progress in text-driven video editing, where the edits are controlled by text only. Tune-a-Video~\cite{tuneavideo} and FateZero~\cite{fatezero} handle video editing by adding additional modules to the state-of-the-art Text-to-Image diffusion models pre-trained on massive image data, and then fine-tune the parameters by training on the target video. Dreamix~\cite{dreamix} proposes a novel mixed fine-tuning model that significantly improves the quality of motion edits. Omer \etal~\cite{text2live} propose Text2Live, which harnesses the richness of information across time, and can perform consistent text-guided editing. In conclusion, all of the above video editing methods are text-guided video-to-video translation, which manipulate the video only through the natural language based on the unprecedented generative power of the Text-to-image diffusion models. However, like the draws of text-driven image editing, due to the natural and inherent gap between the two different modalities of vision and language, it is hard to control the video accurately by relying solely on textual changes. We argue that the language guidance still lacks precise control, whereas additional images reference can better express one’s concrete ideas. As such in this work we proposed a new framework of subject-driven video editing, which uses a reference image as supplemental input to replace cumbersome text descriptions.
\begin{figure*}[t]
  \centering
    \includegraphics[clip,width=\textwidth]{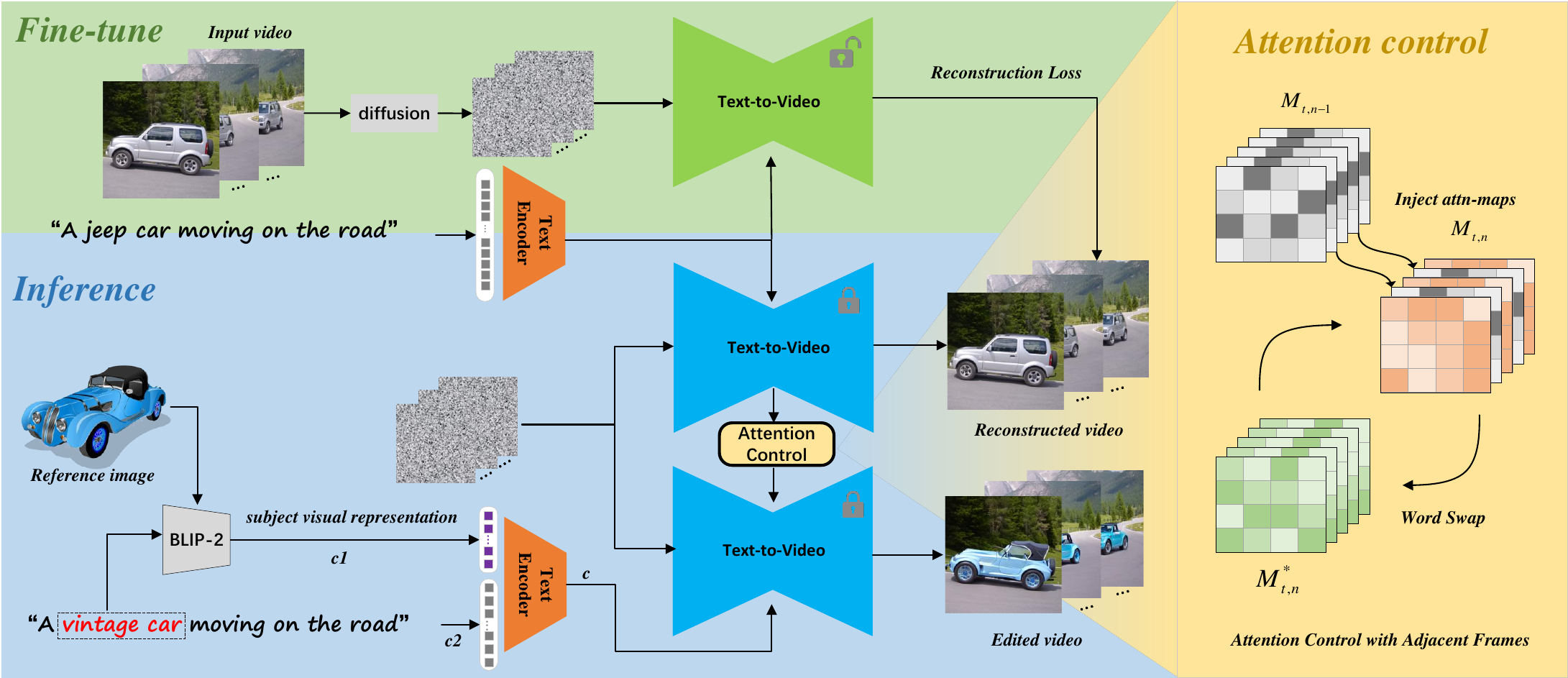}
  \caption{\textbf{Pipeline of \emph{Cut-and-Paste}}. Given a video $\mathcal{V}$ containing $n$ frames which corresponds to the text prompt $\mathcal{P}$ (\eg ``A jeep car moving on the road"), our goal is to edit the input video under the guidance of both the edited prompt $\mathcal{P}^*$ (``A vintage car moving on the road") and the reference image $\mathcal{I}$ (\eg the blue vintage car with black top, at the bottom left of the picture). \textbf{Left}: During fine-tune stage, we update the matrices in attention blocks just like FateZero~\cite{fatezero} to reconstruct the source video. For inference, two already trained 3D-Unet accept different conditional inputs for editing the original video respectively. One accepts the same text prompt as fine-tune progress, another takes the edited prompt ($c1$) and a subject visual representation ($c2$, the output of multimodal encoder BLIP-2) as the input. And finally output the Reconstructed video and Edited video respectively. \textbf{Right}: The attention control with adjacent frames, we inject the attention maps of the previous frame $M_{t,n-1}$ to the current frame $M_{t,n}$ before performing the Word Swap operation in P2P~\cite{prompttoprompt}.} 
  \label{fig:pipeline}
\end{figure*}
\section{Method}
\label{sec:method}
 Given a video $\mathcal{V}$ containing $n$ frames, which corresponds to the text prompt $\mathcal{P}$, our goal is to edit the input video under the guidance of both the edited prompt $\mathcal{P}^*$ and the reference image $\mathcal{I}$, to a new video $\mathcal{V}^*$ with $n$ frames. 
 
In this section, we will first briefly review some preliminaries of diffusion models in Sec.~\ref{sec:Preliminary}, followed by a detailed description of our method in Sec.~\ref{sec:method_subject driven} and Sec.~\ref{sec:method_attn_con}. The pipeline of our approach is depicted in Figure~\ref{fig:pipeline}.
\subsection{Preliminary}
\label{sec:Preliminary}
Diffusion models~\cite{ddim,ddpm,ldm,glide,prompttoprompt,classifierfree} are a family of probabilistic generative models that are trained to learn a data distribution by progressively removing a variable (noise) sampled from an initial Gaussian distribution. With the noise gradually adding to $z$ for $t$ steps, a latent variable $z$ and its noisy version $z_t$ will be obtained, the objective function of latent diffusion models can be simplified to
\begin{equation} \mathbb{E}_{z,c,\epsilon\sim\mathcal{N}(0,1),t} \Bigl[\|\epsilon-\epsilon_{\theta}(z_t,t)\|^2_2\Bigr],
\end{equation}
which is the squared error between the added noise $\epsilon$ and the predicted noise $\epsilon_{\theta}(z_t,t)$ by a neural model $\epsilon_\theta$ at time step $t$, given $c$ as condition. This approach can be generalized for learning a conditional distribution, the network $\epsilon_{\theta}(z_t,t,c)$ can faithfully sample from a distribution conditioned on $c$. In this work, we leverage a pre-trained text-to-image Latent Diffusion Model (LDM), \ie, Stable Diffusion~\cite{ldm}, in which the whole diffusion process is proceeded in the latent space of a pre-trained image autoencoder. To process video, we use the video generation model like FateZero~\cite{fatezero} and Tune-a-Video~\cite{tuneavideo} for their generalization abilities.
\begin{figure*}
  \centering
  \begin{overpic}[width=\textwidth]{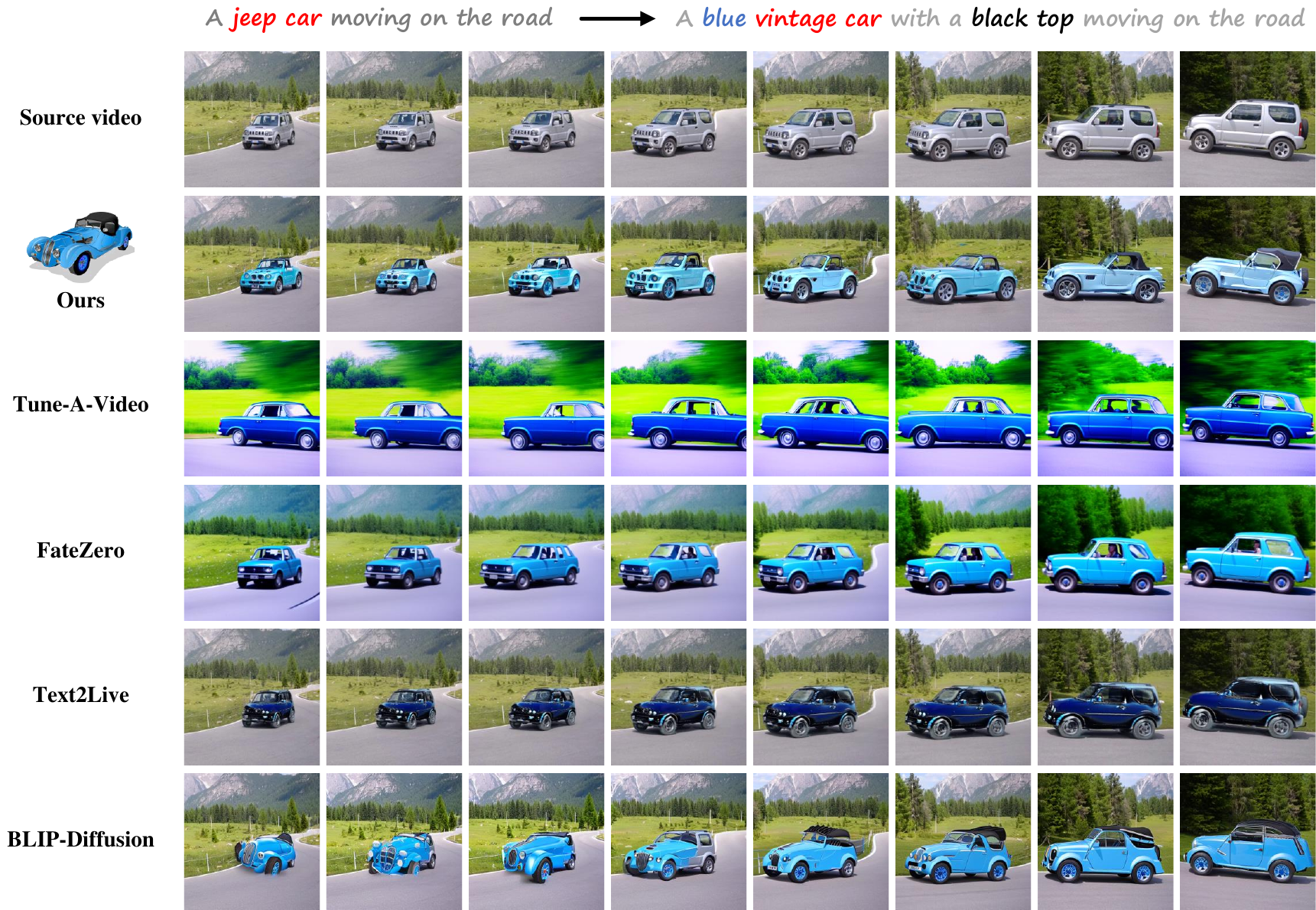} 
    \put(11.5,37.9){\scriptsize \cite{tuneavideo}}
    \put(10,26.8){\scriptsize \cite{fatezero}}
    \put(10.4,15.8){\scriptsize \cite{text2live}}
    \put(11.8,4.7){\scriptsize \cite{blipdiffusion}}
    \end{overpic}
  \caption{\textbf{Visual comparison with \textit{text-driven video editing} and \textit{our subject-driven image editing}.} Our goal is to edit the input video ``A jeep car moving on the road" to ``a blue vintage car with a black top moving on the road", maintaining both video background and temporal consistency while manipulating the edit subject. By the way, for the sake of comparative fairness, we use a relatively rough description for our method and subject-driven image editing, \ie ``A vintage car moving on the road", with no description of the car’s exterior.}
  \label{fig:comparison}
\end{figure*}
\subsection{Subject-Driven Video Editing}
\label{sec:method_subject driven}
For solving the problem of lacking precise control in traditional text-driven video editing model, we aim to propose a novel framework for fine-grained semantic video editing under the guidance of both text prompt and additional reference image. Our first task is to make the reference image an additional input control condition for the traditional text-driven video editing model. To this end, we analyze the text-conditioned model in depth and observe that the key to effect the generation results is the cross-attention mechanism of the UNet backbone~\cite{unet2015}, which is effective for fusing visual and textual embedding. We notice that in order to pre-process the condition $c$ from various modalities (\eg text prompts), a domain specific encoder (\eg text encoder) $\conditioner$ is introduced to project $c$ to an intermediate representation $\conditioner(c) \in \R^{M\times d_\tau}$, which is then mapped to the intermediate layers of the UNet via a cross-attention layer implementing $\text{Attention}(Q, K, V) = M \cdot V$, with
\begin{equation}\label{eq:attention map}
M=\text{Softmax}\left(\frac{QK^T}{\sqrt{d}}\right),
\end{equation}
\begin{equation}
Q = W^{(i)}_Q \cdot  \varphi_i(z_t), \; K = W^{(i)}_K \cdot \conditioner(c),
  \; V = W^{(i)}_V \cdot \conditioner(c) . \nonumber
\end{equation}
Here, $\varphi_i(z_t) \in \R^{N \times d^i_\epsilon}$ denotes a (flattened) intermediate representation of the UNet implementing $\model$ and $W^{(i)}_V \in \R^{d \times d^i_\epsilon}$, $W^{(i)}_Q \in \R^{d \times d_\tau} $ \& $W^{(i)}_K \in \R^{d \times d_\tau}$ are learnable projection matrices, and where $d$ is the latent projection dimension of the keys and queries. Hence, to guide the layout with the reference image, an intuitive strategy is to incorporate the reference image as an extra input control condition in $\conditioner$. Consequently, the pivotal step in problem resolution is to incorporate the representation of the reference image into the textual condition $c$, requiring alignment between the representations of the reference image and the textual prompts. Similar to~\cite{blipdiffusion}, we decide to refine the BLIP-2~\cite{blip2}, a vision-language pre-trained model which successfully produces high-quality text-aligned visual representation, to extract text-aligned subject representation. As visualized in Figure~\ref{fig:pipeline}, the multimodal encoder BLIP-2 takes the reference image and it's corresponding subject words as input and produces a subject visual representation $c1$, which includes an extremely robust correspondence between the reference image and the subject words. Then we concatenate the subject visual representation $c1$ with the text prompt token embeddings $c2$ to jointly serve as input control condition $c$. Finally, like what the general Text-to-Image diffusion model does, the fused control condition $c$ is passed through the CLIP text encoder, serving as the final guidance for the diffusion model to generate the output. 
\setlength{\textfloatsep}{0pt}
\begin{algorithm}[t]
\SetAlgoLined
\textbf{Input:} A source prompt $\mathcal{P}$, a target prompt $\mathcal{P}^*$, and a random seed $s$.\\
\textbf{Output:} Source video $\mathcal{V}$ and edited video $\mathcal{V}^*$.\\
 Latent features from DDIM inversion: $z_{T}$;\\
 $z_{T}^* \gets z_{T}$; \\
 \For{$t=T,T-1,\ldots,1$}{
    $z_{t-1}, M_{t,n-1}, M_{t,n} \gets DM(z_{t},\mathcal{P},t,s)$\;
    $M_{t,n}^* \gets DM(z_{t}^*,\mathcal{P}^*,t, s)$\;
    $M_{t,n}^{+} \gets Inject(M_{t,n-1}, M_{t,n}, t)$\;
    $\widehat{M}_{t,n} \gets Edit(M_{t,n}^{+}, M_{t,n}^*, t)$\;
    $z_{t-1}^* \gets DM(z_{t}^*,\mathcal{P}^*,t,s)\{M_{t,n}^* \gets \widehat{M}_{t,n}\}$\;
 }
 \textbf{Return} $(z_{0},z_{0}^*)$
 \caption{Prompt-to-Prompt frame editing}
 \label{alg:attention}
\end{algorithm}
\subsection{Attention Control with Adjacent Frames}
\label{sec:method_attn_con}
To limit the editing area on real images, existing works~\cite{prompttoprompt,text2live} have proposed the text-based localized editing method without relying on any user-defined mask to signify the editing region. However, it is challenging to achieve such a goal in video editing for maintaining the spatio-temporal consistency across frames. We amend the attention control methods proposed by Prompt-to-Prompt~\cite{prompttoprompt} by inject attentions maps with adjacent frames. As Eq.(\ref{eq:attention map}) shows, the pixel queries $Q$ and token keys $K$ (from condition $c$) are fused to spatial attention maps $M$, and the pixels are more related to the tokens (words) that describe them, \eg , pixels of the jeep-car are correlated with the word "jeep-car". So for frame $n$, if we override the attentions $M_{t,n}^{*}$ that were obtained from the edited prompt $\mathcal{P}^*$, with the $M_{t,n}$ generated by source prompt $\mathcal{P}$, the output frame $n^*$ will be edited by $\mathcal{P}^*$ and meanwhile preserve the structure and background of input frame n. However, processing each frame individually can lead to inconsistencies caused by object motion. Thus, as shown on the right side of Figure~\ref{fig:pipeline}, we inject the attention maps of the previous frame $M_{t,n-1}$ to the current frame $M_{t,n}$ ($Inject(M_{t,n-1}, M_{t,n}, t)$) before performing the Word Swap operation in P2P~\cite{prompttoprompt} in some steps of the diffusion process, which effectively strikes a balance between maintaining video structure and  spatio-temporal consistency across frames. Formally, the pseudo algorithm is shown in Alg.~\ref{alg:attention}. We define the $Edit(M_{t,n-1}, M_{t,n}, M_{t,n}^*, t)$ to be the edit function described above, where the function $DM$ means overriding the attention map $M$ with an additional given map $\widehat{M}$, but keep the values $V$ from the supplied prompt. Our extensive experiments further demonstrate that this mechanism can achieve better editing results.
\begin{figure*}
  \centering
    \includegraphics[clip,width=\textwidth]{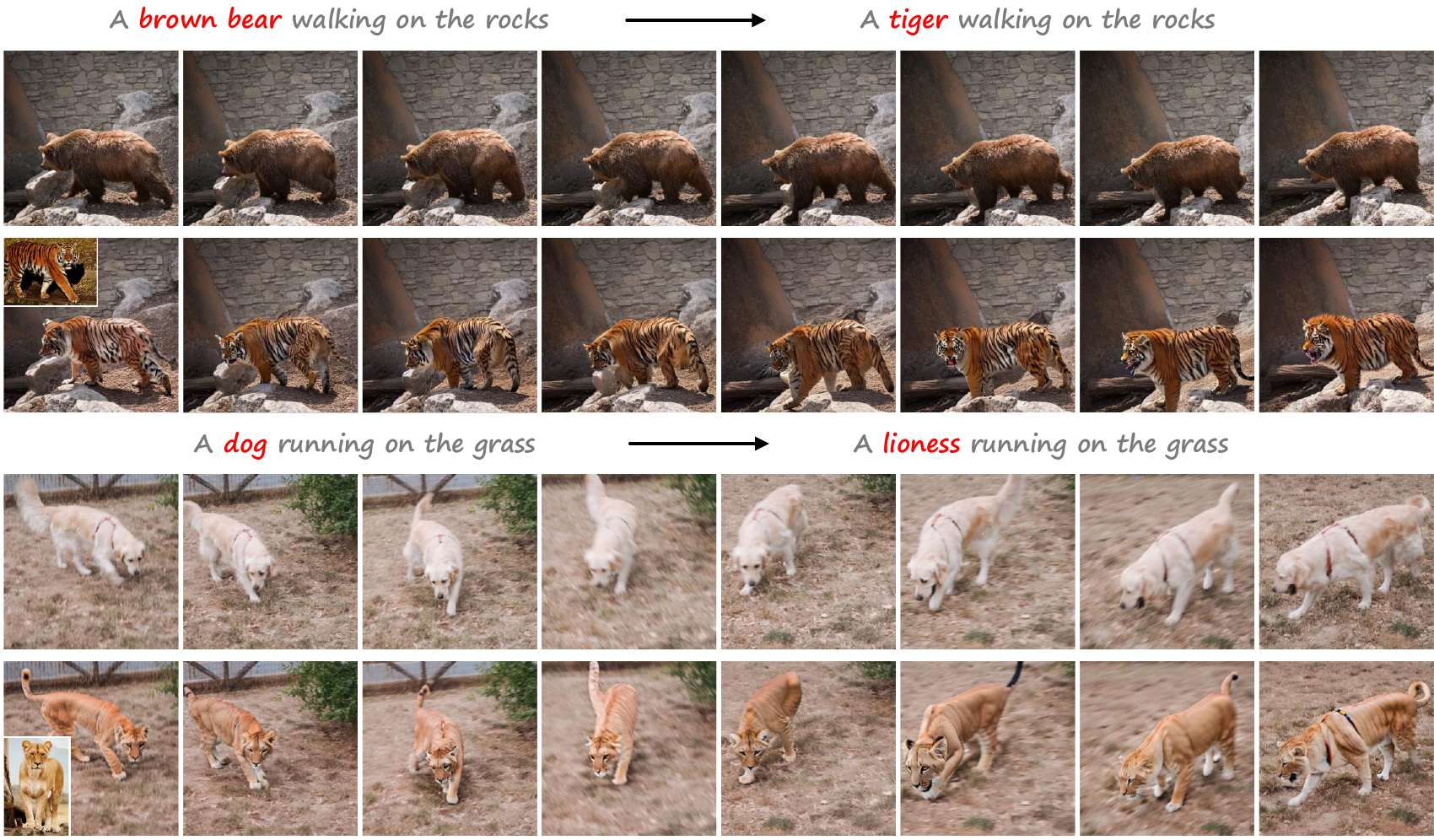}
  \caption{\textbf{Performance of our \emph{Cut-and-Paste}.} Our approach achieves fine-grained semantic video editing and preserves the background and maintains spatio-temporal consistency of the original video. By the way, all the text prompts in the experiment of our method are simple, without any words to describe the object’s properties.}
  \label{fig:performance}
\end{figure*}
\begin{figure*}
  \centering
 \includegraphics[clip,width=\textwidth]{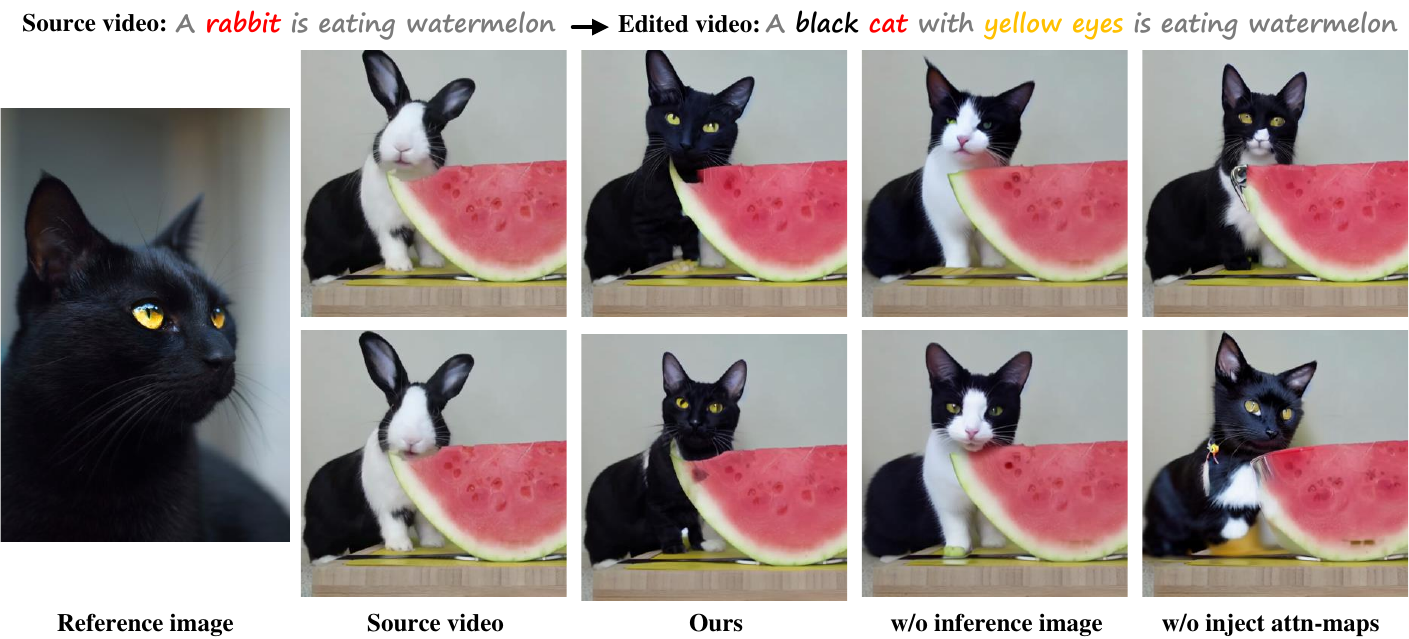}
  \caption{\textbf{Ablation study.} We study the effects of removing the supplementary input of reference image and the inject attention maps components. We can find that w/o inference image, the appearance of the generated cat is black and right, similar to the rabbit in source video, but not match the text prompts (For all experiments without reference image as supplementary input, we use a more detailed textual description as prompt input). Also, w/o inject attn-maps, the generated results exhibit large variations between frames.}
  \label{fig:ablation}
\end{figure*}
\section{Experiments}
\label{sec:experiments}
\subsection{Implementation Details}
In order to manipulate the real-word videos, we apply our method on several videos from DAVIS~\cite{davis2017} and other in-the-wild videos to evaluate our approach. Similar to Tune-a-Video~\cite{tuneavideo} and FateZero~\cite{fatezero}, we fix the image autoencoder and sample 8 or 24 frames at the resolution of 512 $\times$ 512 from a video. The source prompt of the video is generated via the image caption model Blip-2~\cite{blip2}. We design the target prompt for each video by replacing the subject words and apply additional reference images as a supplementary input, which all of the reference images are collected from the Web and will be published in the future. The DDIM~\cite{ddim} sampler is set to 50 steps. During attention control, we set the cross-attention replacing ratio to 0.8 and the attention threshold to 0.3. Finally, approximately 5 minutes are required for fine-tuning, and around 1 minute for inference, for a single video on a single NVIDIA A100 GPU, which is comparable to general text-driven video editing methods.
\subsection{Visual Comparison}
\setlength{\textfloatsep}{7pt}
\begin{figure}[t]
  \centering
\includegraphics[clip,width=\linewidth]{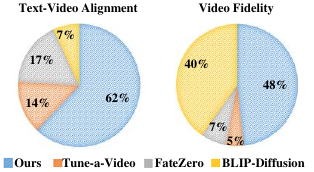}
  \caption{\textbf{User study on text-video alignment and video fidelity.} }
  \label{fig:userstudy}
\end{figure}
We compare our editing results primarily with the two main types of semantic editing methods: 1) \textit{text-driven video editing}, which manipulates the video only under the guidance of text prompt, including Tune-a-Video~\cite{tuneavideo}, FateZero~\cite{fatezero} and Text2Live~\cite{text2live}; 2) \textit{subject-driven image editing}, which edits the image guided by both text prompt and a reference image, including BLIP-Diffusion~\cite{blipdiffusion}. 

\textbf{(1) Comparison to text-driven video editing.}
For the sake of comparative fairness, since our method uses an additional reference image as supplementary input, which contains fine-grained semantic information. So for the same source video ``A jeep car moving on the road", we use a more detailed textual description as the edited prompt to guide the text-driven video editing model, \ie, ``A blue vintage car with a black top moving on the road", whereas a relatively rough description for our method like ``A vintage car moving on the road", with no description of the car's exterior. The qualitative comparison results are displayed in the first to penultimate rows of Figure~\ref{fig:performance}. \textbf{i)} \textbf{Tune-a-Video}~\cite{tuneavideo} struggles to preserve the appearance and structure of the original video, which can only make out a blue car moving on the road. \textbf{ii)} \textbf{FateZero}~\cite{fatezero} performs well in terms of the temporal consistency, however, the edited results do not exactly match the target prompt, missing the feature of ``black top". Also, compared to the original video, the output video has an overall bluish hue and doesn't hold the background very well. We speculate the reason is that the text-based model lacks control over specific semantic regions. \textbf{iii)} \textbf{Text2Live}~\cite{text2live} achieves great effects in maintaining both video background and temporal consistency, while due to its reliance on layered neural atlases, it can not change the shape to match the target object, \eg from ``jeep car" to ``vintage car". And the body color mixes black and blue which not reconstructs well. As can be seen, our method \textbf{Cut-and-Paste} (the second row in Figure~\ref{fig:performance}) enables fine-grained control over the generated structure and exhibits high fidelity to the structure and scene layout of source video. More performance can be seen in Figure~\ref{fig:performance}.
\indent \textbf{(2) Comparison to subject-driven image editing.}
For the comparison with subject-driven image editing, we use the same and simple edited prompt and reference image as the condition. As shown in Figure~\ref{fig:performance} (last row), We can find that the editing results are terrible when directly applying the method of BLIP-Diffusion~\cite{blipdiffusion} to the video editing. Only part of the frames show satisfactory results. We guess that it is the result of a lack of fine-tuning on the original video and a failure to utilize the features across frames. In contrast, Since our method is appropriately fine-tuned before inference and utilizes features between neighboring frames in attention control, the motion variations between each frame are greatly reduced.
\begin{table}[t]
\centering
\resizebox{\columnwidth}{!}{
\large
\renewcommand\arraystretch{1.5}
  \begin{tabular}{c|c|ccc}
    \toprule
    Method & Type & CLIP Score $\uparrow$ & LPIPS-P $\downarrow$ \\
    \hline
    Cut-and-Paste (ours) & subject-driven video editing & \textbf{0.7089} & \textbf{0.2112}\\
    \hline
    Tune-A-Video~\cite{tuneavideo} & \multirow{2}{*}{text-driven video editing} & 0.6780 & 0.6376\\
    FateZero~\cite{fatezero} &  & 0.6442 & 0.4472 \\
    \hline
    BLIP-Diffusion~\cite{blipdiffusion} & subject-driven image editing & 0.6901 & 0.2295 \\
    \bottomrule
  \end{tabular}
}
\caption{\textbf{Quantitative comparison with different methods.} We evaluate the text-image similarity through CLIP Score, and the spatio-temporal consistency through LPIPS.}
\label{table:cpmparison}
\end{table}
\subsection{Quantitative Evaluation}
We numerically evaluate the results according to these complementary metrics: 1) CLIP Score~\cite{clipscore}, the text-image similarity to quantify how well the edited video comply with the text prompt (higher is better) 2) LPIPS~\cite{lpips}, deviation from the original video frames (lower is better). 

The quantitative results are summarized in Table~\ref{table:cpmparison} (note that Text2Live was not evaluated due to the absence of pre-trained models on other datasets). We can find that our results get a higher CLIP Score and a lower LPIPS, whether comparing to the methods of text-driven video editing or subject-driven image editing, which demonstrates that, not only can our method generate videos that align well with textual descriptions, but also achieves a better trade-off between preserving the structure of original video and spatio-temporal consistency across frames.
\textbf{User study.} In order to obtain the user's subjective evaluation of the edited video, we conduct a user study on 87 participants who are mainly students in university. For text-video alignment, we present the text prompt and the videos generated by different methods and ask ``which video aligns with the textual description better?" For video fidelity, we present the original video and generated side by side and ask ``which video preservers the background and temporal consistency of the original video better?" The results of the evaluation can be seen in Figure~\ref{fig:userstudy}. As can be seen, users exhibit a strong preference towards our method both on the metric of text-video faithfulness and video fidelity.
\begin{figure}
  \centering
\includegraphics[clip,width=\linewidth]{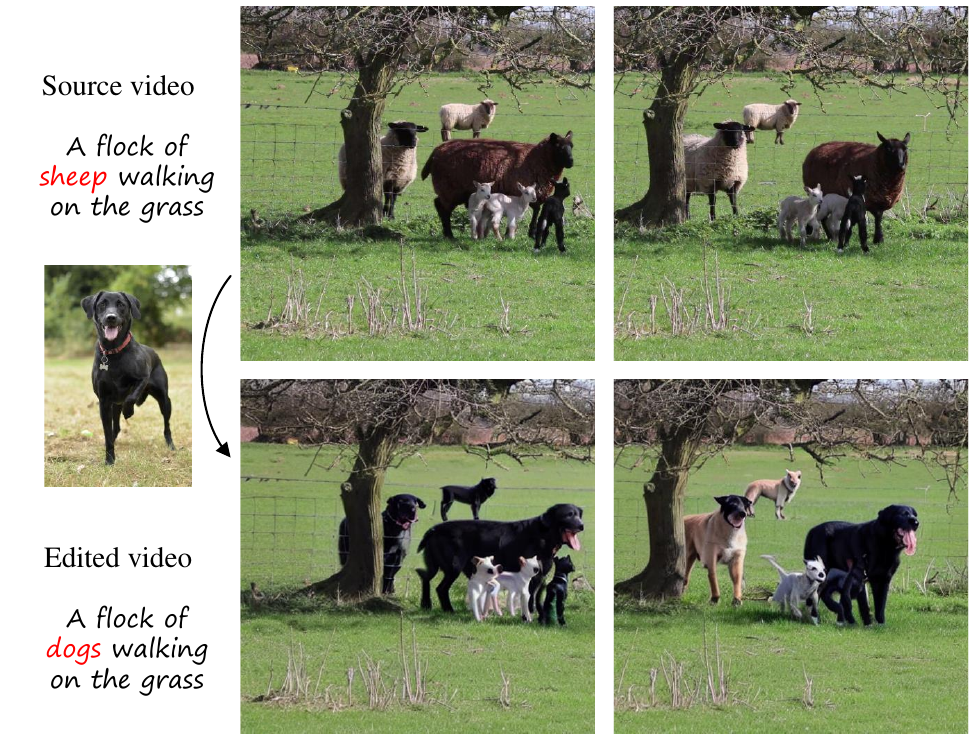}
  \caption{\textbf{Failure case 1.} Our model is not capable of editing multiple objects at the same time, \eg when changing the video ``A flock of sheep walking on the grass" to ``A flock of dogs walking on the grass" with a image of dog as supplementary input, not all sheep have been converted to dogs. }
  \label{fig:limitation1}
\end{figure}

\begin{figure}
  \centering
\includegraphics[clip,width=\linewidth]{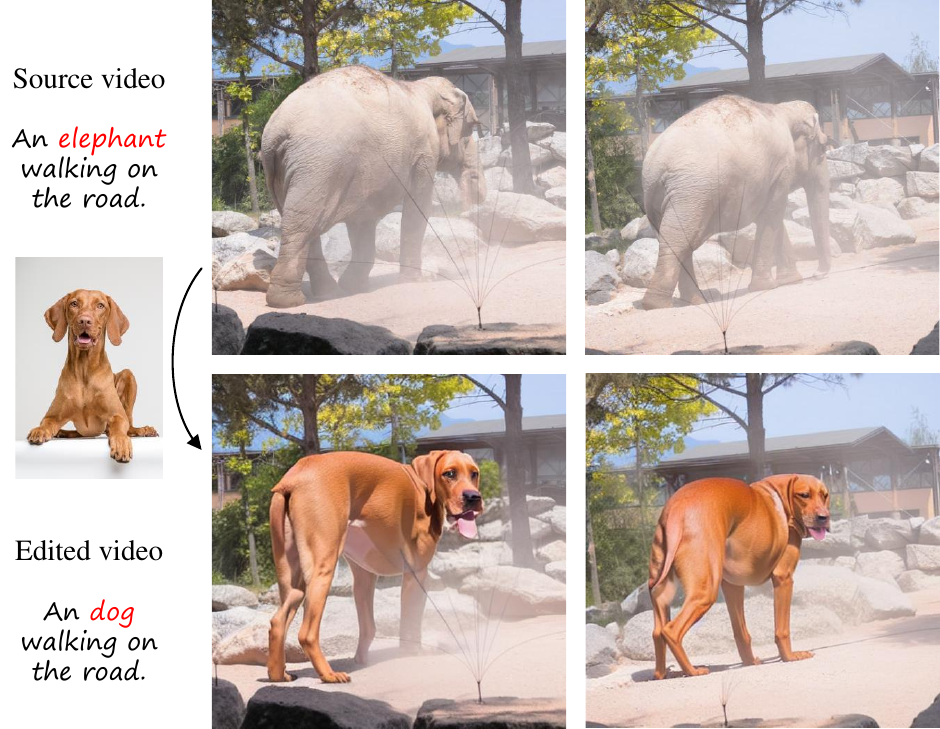}
  \caption{\textbf{Failure case 2.} Our model struggles to change the size of the two editing objects on a large scale, \eg changing the video ``An elephant walking on the road" to ``A dog walking on the road" with a image of dog as supplementary input. Even though our method can effectively capture the details of the dog, the large difference in body size between the two resulted in undesirable editing results, which looks strange.}
  \label{fig:limitation2}
\end{figure}
\subsection{Ablation Study}
\label{sec:ablation}
We ablate our key design choices by evaluating the performance in the following cases: (i) w/o introducing the reference image as a complementary input (w/o inference image), (ii) w/o attention control with adjacent frames (w/o inject attn-maps). The performance is shown in Figure~\ref{fig:ablation}. We can find that w/o inference image, the appearance of the generated cat is black and right, similar to the rabbit in source video, but not match the text prompts. Also, w/o inject attn-maps, the generated results exhibit large variations between frames. The results demonstrate that both complementary input of reference image and attention control with adjacent frames are critical for fine-grained semantic video editing -- the reference image provides more fine-grained information than plain text, while attention control with adjacent frames achieves a better balance preserving the structure and spatio-temporal consistency of the original video.
\section{Limitations and Future Works}
\label{sec:Limitations and Future Works}
While we have demonstrated that our approach is able to provide more precise control and fine-grained semantic generation for localized video editing, there still exist a number of limitations. Firstly, our model struggles to manipulate multiple objects at the same time, which is constrained by the capabilities of the fundamental T2I diffusion model. Secondly, since we locate the editing region just by leveraging the correspondence between pixels and words in attentions maps, which successfully avoids masking the editing area manually, while brings up another issue -- struggle to change the size of the editing object on a large scale. Finally, we find that our current method is challenging in removing existing object in each frame just like P2P and general text-driven video editing methods. Failure cases are shown in Figure~\ref{fig:limitation1} and Figure~\ref{fig:limitation2}. Our future work may focus on further expanding the applicability of our approach and removing the fine-tune process to make it more convenient.
\section{Conclusion}
\label{sec:conclusion}
In this work, we present Cut-and-Paste, a subject-driven video editing method, which is a novel framework for real-world semantic video editing guided by a plain text prompt and an additional reference image. We firstly introduce the reference image as supplementary input to general text-driven video editing, without racking your brain to come up with a text prompt describing the detailed appearance of the object. Besides, the design of attention control with adjacent frames achieves a better balance preserving the background and spatio-temporal consistency of the original video. We conduct extensive experiments and demonstrate the superior qualitative and quantitative results of our model, compared to the state-of-the-art methods of both text-driven video editing and subject-driven image editing.
\label{sec:acknowledgements}

\textbf{Acknowledgements:}
This work is supported by the National Natural Science Foundation of China (62072151, 72004174, 61932009,
62020106007, 62072246), Anhui Provincial Natural Science
Fund for the Distinguished Young Scholars (2008085J30),
Open Foundation of Yunnan Key Laboratory of Software Engineering (2023SE103), CCF-Baidu Open Fund and CAAI-Huawei MindSpore Open Fund. Corresponding author: Zhao Zhang.  
{
    \small
    \bibliographystyle{ieeenat_fullname}
    \bibliography{main}
}
\end{document}